\documentclass[conference]{IEEEtran}
\usepackage{amsmath,amssymb,amsfonts}
\usepackage{hyperref}
\hypersetup{colorlinks,allcolors=black}
\usepackage{graphicx}
\usepackage{stfloats}
\usepackage{textcomp}
\usepackage{threeparttable}
\usepackage{xcolor}
\usepackage{booktabs}
\usepackage{multirow}
\usepackage{listings}
\usepackage{algorithm}   % Floating environment for algorithms
\usepackage{algpseudocode} % Specific algorithmic environment (pseudocode style)
\usepackage{setspace} 
\usepackage{siunitx}
\AtBeginEnvironment{algorithmic}{\setstretch{1.2}} 
\def\BibTeX{{\rm B\kern-.05em{\sc i\kern-.025em b}\kern-.08em
    T\kern-.1667em\lower.7ex\hbox{E}\kern-.125emX}}
\begin{document}
\bstctlcite{IEEEexample:BSTcontrol}

\title{Moment and Highlight Detection via MLLM Frame Segmentation\\}

\author{\IEEEauthorblockN{I Putu Andika Bagas Jiwanta}
\IEEEauthorblockA{\textit{School of Electrical Engineering and Informatics} \\
\textit{Institut Teknologi Bandung}\\
Bandung, Indonesia \\
bagasjiwanta@gmail.com}
\and
\IEEEauthorblockN{Ayu Purwarianti}
\IEEEauthorblockA{\textit{School of Electrical Engineering and Informatics} \\
\textit{Institut Teknologi Bandung}\\
Bandung, Indonesia \\
ayu@staff.stei.itb.ac.id}
}

\maketitle

\begin{abstract}
Detecting video moments and highlights from natural-language queries have been unified by transformer-based methods. Other works use generative Multimodal LLM (MLLM) to predict moments and/or highlights as text timestamps, utilizing its reasoning capability. While effective, text-based generation cannot provide direct gradients for frame-level predictions because the model only emits language tokens. Although recent Reinforcement Learning (RL) methods attempt to address the issue, we propose a novel approach by applying segmentation objectives directly on the LLM's output tokens. The LLM is fed with a fixed number of frames alongside a prompt that enforces it to output a sequence of continuous ``0'' and/or ``1'' characters, with one character per frame. The ``0''/``1'' characters benefit from the LLM's inherent language capability while also acting as background and foreground probabilities, respectively. Training employs segmentation losses on the probabilities alongside a normal causal LM loss. At inference, beam search generates sequence and logits, acting as moments and saliency scores, respectively. Despite sampling only 25 frames---less than half of comparable methods---our method achieved strong highlight detection (56.74 HIT@1) on QVHighlights. Additionally, our efficient method scores above the baseline (35.28 MAP) for moment retrieval. Empirically, segmentation losses provide a stable complementary learning signal even when the causal LM loss plateaus.

% Code and training details are available at \url{https://github.com/bagasjiwanta/moment-retrieval-frame-segmentation}
\end{abstract}

\begin{IEEEkeywords}
moment retrieval, highlight detection, segmentation, LLM
\end{IEEEkeywords}

\section{Introduction}
Video moment retrieval (MR) aims to localize temporal boundaries of a video activity given a natural-language query~\cite{qvhighlights_2021,zhang2022actionformer}. Highlight detection (HD) identifies the most salient clips for the query~\cite{qvhighlights_2021}. Typically, unified MR and HD methods~\cite{qvhighlights_2021,liu2022umt,moon2023query} predict the two tasks by adopting custom prediction heads on top of a Transformer~\cite{vaswani2017attention}. Since evaluation metrics for MR are based on IoU, some use differentiable IoU variants such as GIoU~\cite{rezatofighi2019generalized}, while others reframe the task. In general, they meticulously craft their prediction head and/or attention design for the task. 

Recent advancements use MLLM that pair a visual encoder with an LLM, such as Chrono~\cite{Meinardus_2025_ICCV}, LLaVA-MR~\cite{llava_mr_2024}, and VTG-LLM\cite{vtg_llm_2024}. These models provide a simple design by framing the problem as next token prediction, generating timestamps as text. However, this design disconnects the training objectives from frame-level visual signals, since standard LM training objectives cannot differentiate IoU-based loss. Furthermore, even if IoU can be computed in the forward pass, backpropagation is still blocked for text-formatted spans because the decoding step is non-differentiable. To bridge this gap, recent approaches introduced RL to take advantage of the non-differentiable metrics \cite{tempo_r0_2025,li2025tempsampr1effectivetemporalsampling}. 

The segmentation task has been shown to benefit from well-designed objective and combinations of several objectives \cite{abraham2019focaltversky,taghanaki2019combo,azad2023loss}. However, semantic segmentation with MLLMs has only been explored recently in~\cite{lisa_2024,jang2025mmr,qian2025reasoning}, where embeddings of a specific \texttt{<SEG>} token is treated as a segmentation mask. In addition, others proposed next token prediction methods because it is deemed superior as it does not lose conversational ability~\cite{wu2024visionllm,liu2025seg,lan2024text4seg}. However, no prior work has applied segmentation objectives specifically on the output tokens for MR and HD. 

This paper proposes frame segmentation modelling for both tasks, using the BLIP-3 MLLM~\cite{blip3_2025}. Instead of generating timestamps directly, the model outputs binary textual masks over frames to indicate the relevant moments. Adapting to the dataset characteristics, the objective is a summation of binary cross-entropy loss, Tversky loss\cite{tversky_2017}, generalized dice loss\cite{generalized_dice_2017}, and the standard causal LM loss. To conclude, the contributions of this paper are:
\begin{enumerate}
    \item Propose a new segmentation method for moment retrieval and highlight detection with MLLM, where segmentation objectives are directly applied on the output tokens alongside a causal LM objective.
    \item Demonstrate that the method allows efficient learning on highlight detection, while also producing acceptable results on moment retrieval. We achieve this using a frame count of 25, less than half of similar techniques, establishing a strong foundation for future scaling.
\end{enumerate}

\footnotetext{Code and implementation details: \url{https://github.com/bagasjiwanta/llm-frame-segmentation}}

\section{Related Works}

\subsection{Traditional Joint Moment Retrieval and Highlight Detection Methods}
Moment-DETR~\cite{qvhighlights_2021} pioneered the union by both predicting highlights and moments from the same encoder-decoder Transformer. Technically, it acts as a direct set prediction MR rather than a proposal-based one via the usage of $n$-sized learned latent queries, where $n$ is determined beforehand. The decoder output $E_{dec}\in{\mathbb{R}^{n\times d_{model}}}$ is projected by a prediction head to $\mathbb{R}^{n\times 2}$, containing the center and width of each moment, respectively. Aimed to truly unify MR and HD, UMT \cite{liu2022umt} eliminated the hardcoded prediction number $n$. It instead predicts both moments and highlights from the decoder output $E_{dec}\in{\mathbb{R}^{f}}$, where $f$ is the frame count. Although unification is pushed further in QD-DETR\cite{moon2023query}, TR-DETR~\cite{sun2024tr}, and UniVTG~\cite{lin2023univtg}, they lack semantic reasoning and only rely on task-specific prediction heads.

\subsection{Large Language Models for Temporal Localization}

Advances in multimodal LLMs for language-vision understanding have shifted recent works towards generative approaches. MR-only methods such as Chrono~\cite{Meinardus_2025_ICCV} and LlaVA-MR~\cite{llava_mr_2024} generate spans as timestamp texts in $[[ts^{1}_{start}, ts^{1}_{end}],[ts^{2}_{start}, ts^{2}_{end}],...]$ format. Methods supporting both tasks show limited performance in HD (see Table~\ref{tab5:qvh_test_set_performance}). Improvements for the setup are limited to how well the model learns prompt, formatting, and language labels instead of directly improving via region-level or frame-level objective. 

Few recent works have introduced RL to solve the non-differentiable nature of evaluating text-generated timestamps. Tempo-R0~\cite{tempo_r0_2025} introduced an variation of the standard GRPO~\cite{shao2024deepseekmathpushinglimitsmathematical} by introducing Partial Irrelevance Refusing-based GRPO (PIR-GRPO). It reinforces the model's ability to reject incorrect moments and refine temporal boundaries. TempSamp-R1~\cite{li2025tempsampr1effectivetemporalsampling} addresses limitations in GRPO by integrating off-policy guidance to reduce the large temporal search space. The RL methods are effective but needed well-designed reward and value.

\section{Method}

\subsection{Method Overview}
The goal is to design a framework to detect both moments and highlights using LLM generation output while also allowing it to learn from segmentation objectives. To achieve this, a video $\mathbf{V}\in{\mathbb{R}^{n_v\times c\times h\times w}}$ is divided to $f$ clips by sampling the middle frame for each, where $n_v$ is the number of frames in the video. Subsequently, these frames are passed through a vision encoder and tokenizer to obtain vision tokens $\mathbf{V}^\prime\in{\mathbb{R}^{f\times l\times d_{model}}}$ where $l$ is the number of tokens attributed per frame. 

\begin{figure*}[ht]
    \centering
    \includegraphics[width=1.0\linewidth]{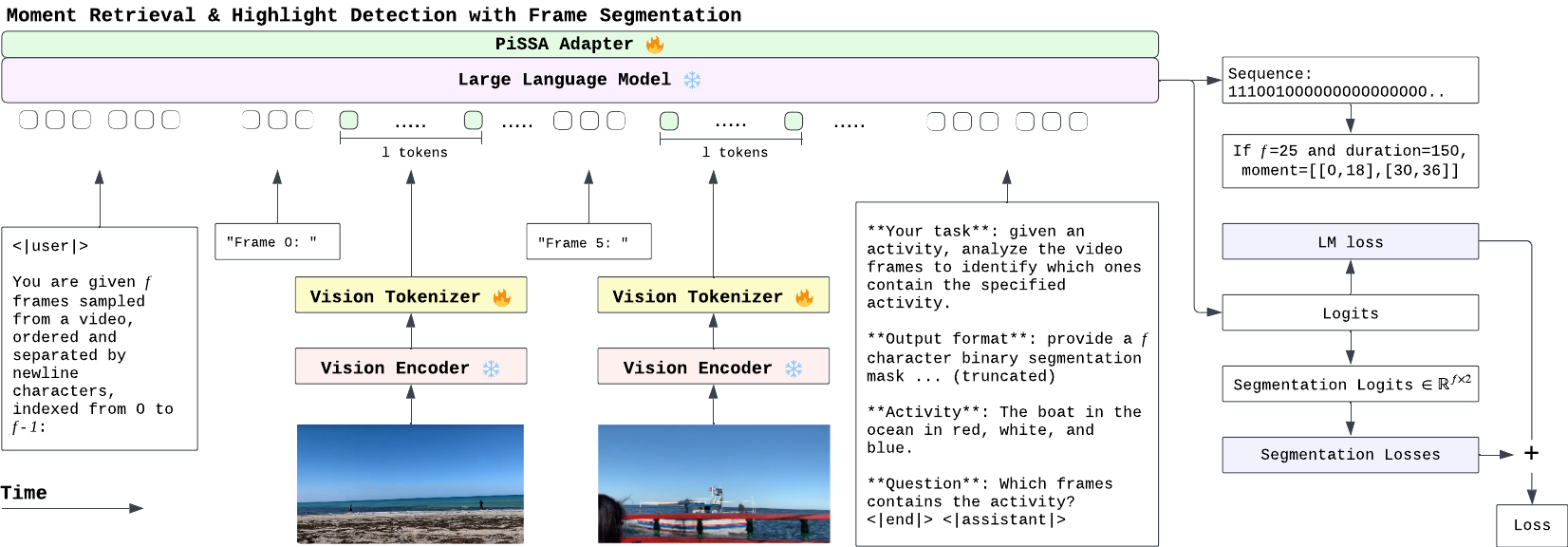}
    \caption{\textbf{Moment retrieval and highlight detection with frame segmentation.} The model processes interleaved visual-text input, where each frame is projected to $l=128$ visual tokens. We use instruction tuning with a specific activity query to generate binary segmentation mask as \texttt{0}/\texttt{1} text. Moments are converted from beam search decoding output, while saliency scores are derived from the segmentation logits. The model is optimized using a joint objective by summing the language modeling and segmentation losses.}
    \label{fig:model_architecture}
\end{figure*}

The model is then tasked to output $f$ continuous \texttt{0} and/or \texttt{1} tokens. For each of the $f$ output character, we extract its logits by taking vocabulary-logit values corresponding to the token IDs of \texttt{0} and \texttt{1}. The probability of the foreground is computed as $\text{softmax}(\mathbf{z})[1]$ (zero-indexing) where $\mathbf{z} = [logit\_0, logit\_1]$, while background uses the 0\textsuperscript{th} index. This is then used to compute both segmentation losses and final saliency scores for highlight detection, while the output sequence is used for moment retrieval. Model pipeline can be seen in Figure~\ref{fig:model_architecture}.

\subsection{Segmentation Modelling}
To obtain frame-level predictions from a video with $t$ duration, the MLLM is given $f$ frames and prompted to output $f$ characters. Because the LLM has been pre-trained on large-scale textual and visual data, the segmentation tokens should best utilize the LLM's existing knowledge. Therefore, the characters \texttt{0} and \texttt{1} are used as background and foreground indicators, which the LLM already recognize as negative and positive, respectively. We verify that both \texttt{0} and \texttt{1} are single tokens in the tokenizer used by the BLIP-3 model and have token IDs of 29900 and 29896, respectively. Empirically, the tokenizer treats the binary sequence as exactly $f$ discrete tokens, with no grouping of consecutive characters. 

This setup ensures a one-to-one mapping, with logits corresponding to each token are interpreted as both frame-level predictions and window-level predictions. However, because the model only "sees" one frame of the window, we assign the score strictly to that frame's position rather than averaging across the window. Consequently, this approach heavily relies on the representing frames, meaning that increasing the frame count is the primary method to ensure a better frame-window connection. A simple uniform sampling is used to assign each frame a window of $\frac{t}{f}$ seconds. The sampled frame indices $\mathbf{v}=\{v_0,v_1,...,v_{f-1}\}$ are calculated from the $fps$, $t$, and $f$, where each index $v_i$ is 
\[v_i=\frac{t}{f}\times fps\times (0.5 + i)\]

\subsection{Loss Functions}
The design of the loss functions must be adaptable to the dataset's characteristics. Binary cross-entropy loss ($\mathcal{L}_{bce}$) is used as frame-level loss. By setting the weight for the positives, it is also able to combat data imbalance where background frames appear more often. To improve general shape of the segments, region-level losses are also used: Tversky loss~\cite{tversky_2017} and generalized dice loss~\cite{generalized_dice_2017}. 

Tversky loss ($\mathcal{L}_{tv}$)~\cite{tversky_2017} is used to adjust the importance of false negatives or false positives by adjusting the $\beta$ and $\alpha$ parameter. Meanwhile, generalized dice loss ($\mathcal{L}_{gd}$)~\cite{generalized_dice_2017} is used to accommodate intra-batch imbalances (hard and easy segments simultaneously within a batch) automatically. With the causal LM loss ($\mathcal{L}_{lm}$), the final loss is a weighted sum:
\[    
\mathcal{L} = w_{lm}\mathcal{L}_{lm} + w_{bce}\mathcal{L}_{bce} + w_{tv}\mathcal{L}_{tv} + w_{gd}\mathcal{L}_{gd}
\] where $w_{lm}$, $w_{bce}$, $w_{tv}$, and $w_{gd}$ are scalar weight of each loss components and they are scheduled during training.

\subsection{Training Algorithm}
The overall training process is summarized in Algorithm~\ref{alg:combined_training}, which simplifies the data processing and loading. The process starts by loading the batch and doing a forward pass with the model. From the logits output, segmentation logits is grabbed by indexing the two characters on only the frame positions, and normalized with the softmax function. Segmentation objectives are then calculated from the normalizes logits and the model is updated by the summation of segmentation and causal LM loss. 

\begin{algorithm}
  \caption{Frame segmentation fine-tuning process for a single batch.}
  \label{alg:combined_training}
  \begin{algorithmic}[1]
    \Require Model $M$, Optimizer $\mathcal{O}$, frame count $f$, model sequence length $s$, batch size $b$, loss weights $w_{lm}, w_{bce}, w_{tv}, w_{gd}$, and model vocabulary size $v$.
    \Require Batch $\mathcal{B}=\{\mathbf{X}, \mathbf{Y}_{true}\}$, where $\mathbf{X}\in\mathbb{R}^{b\times s}$ are the input IDs and $\mathbf{Y}_{true}\in\mathbb{R}^{b\times f\times 2}$ are the one-hot encoded segmentation labels.
    \State $(\mathcal{L}_{lm}, \hat{\mathbf{Y}}\in{\mathbb{R}^{b\times s\times v}}) \gets M(\mathbf{X})$ 
    
    \State $\hat{\mathbf{Y}} \in \mathbb{R}^{b \times f \times 2} \gets \hat{\mathbf{Y}}[\colon,0:f, [token\_id\_0,token\_id\_1]]$ 
    \State $\hat{\mathbf{Y}}\gets Softmax(\hat{\mathbf{Y}}, dim=-1)$
    
    \State Compute $\mathcal{L}_{bce}, \mathcal{L}_{tv}, \mathcal{L}_{gd}$ using $\hat{\mathbf{Y}}$ and $\mathbf{Y}_{true}$.
    
    \State $\mathcal{L} \gets w_{lm}\mathcal{L}_{lm} 
                      + w_{bce}\mathcal{L}_{bce} 
                      + w_{tv}\mathcal{L}_{tv} 
                      + w_{gd}\mathcal{L}_{gd}$
    
    \State Compute gradients of $\mathcal{L}$ w.r.t. parameters of $M$.
    
    \State Update parameters of $M$ using optimizer $\mathcal{O}$.
  \end{algorithmic}
  \vspace{1mm}
  \footnotesize\textit{Token IDs 29900 and 29896 correspond to '0' and '1' in the BLIP-3 tokenizer. Because BLIP-3's LLM is left padded, frame positions are always in 0 to f.}
\end{algorithm}

\subsection{Model Inference}
For moment retrieval, we calculate step duration of each frame and convert the generated sequence to moment boundaries. For highlight detection, we first obtain the same $\hat{\mathbf{Y}}\in{\mathbb{R}^{b\times f \times 2}}$ found in Algorithm~\ref{alg:combined_training} at step 3. Then, we calculate effective step duration for each frame to accommodate videos where the duration is not an exact multiple of the sampling interval. To obtain a score for each second, we linearly interpolate between adjacent frame scores according to their temporal position. 
\section{Experimental Setup}

\subsection{Model Setup}

The BLIP-3 Large Multimodal Model (LMM)~\cite{blip3_2025} is chosen as the video LLM, consisting of a SigLIP~\cite{zhai2023sigmoidloss} visual encoder, a Perceiver Resampler~\cite{alayrac2022flamingo} vision tokenizer, and a Phi-3 LLM~\cite{abdin2024phi3technicalreporthighly}. Each frame is processed by the visual encoder and vision tokenizer before being passed to the LLM. The \textit{instruct-interleave} variant with 4.36 billion parameters is used, which has not been trained with video data. During training, following \cite{blip3_2025}, only the visual encoder is kept frozen, since the dataset does not introduce any new image distribution. Figure~\ref{fig:model_architecture} shows how the visual encoder and tokenizer is used to interleave frames into the prompt, and the whole pipeline.

\subsection{Dataset}
We evaluate on the QVHighlights dataset because it is the only benchmark that provides unified annotations for both tasks. Other datasets, such as Charades-STA~\cite{gao2017talltemporalactivitylocalization}, THUMOS 15~\cite{IDREES20171}, and HACS~\cite{zhao2019hacs} only support MR due to the lack of saliency scores. Other datasets with scores such as ActivityNet Captions~\cite{7298698} and EGO4d~\cite{grauman2022ego4d} only have single score per moment, which may not be accurate since saliency within moments can naturally fluctuate. 

The QVHighlights dataset contains 7,218 training samples, 1550 validation samples, and 1542 test samples. More than 95\% of videos have a 150-second duration, with a single query per video. Each video is annotated with soft labels at 2 second intervals, indicating the relevancy of the clip to the query. Parts without labels are assumed to be non-relevant, having 0 score. The dataset shows a significant class imbalance on both train and validation set, with 2.3378 to 1 background to foreground ratio, or 70.04\% background portion. 

Our preprocessing starts by uniformly sampling each video to 25 frames. Then, they are interleaved into the main instruction prompt where each frame will be represented by 128 tokens. Given the small 4096-token context window, 25 frames allow for enough visual information while leaving 896 tokens for detailed textual instructions. To increase data diversity, one left and one right neighboring frame are sampled for each timestamp. The two frames are exactly 3 frames apart from the middle frame to allow for some difference without losing much visual information (shown in Figure~\ref{fig:frame_sampling}).

\subsection{Training Configuration}
To match the class imbalance ratio mentioned at section IV.B, a positive weight of 2.3378 is used for the binary cross-entropy loss. For Tversky loss, 0.7 is set for the $\beta$ parameter, favoring recall, matching the background portion. The weights of the segmentation losses are set equal to each other throughout training. During training, outputs are teacher-forced to 26 tokens (25 scores and 1 end-of-sentence token).

Since BLIP-3 instruct-interleaved variation is not trained with videos, we fully trained the vision tokenizer (110 M parameters). SFT-based models shows that non-LLM part of the model may not be trained \cite{Meinardus_2025_ICCV,llava_mr_2024} and we left this for future work. We implement parameter-efficient fine-tuning to the LLM with PiSSA~\cite{pissa_2024} and rsLoRA~\cite{rslora_2023} and train 135 million parameters in total. PiSSA is implemented on all the linear layers of the LLM, and each layer is updated with rsLoRA's adjustment.

A cosine-decay learning rate scheduler with six warm-up epochs is set. In the warm-up period, the segmentation losses initially have 0 weights and are linearly increased together to let the model learn from causal LM loss first. The complete set of hyperparameters is listed in Table~\ref{tab:hyperparams}. Training for 11 epochs (excluding evaluation time) completed in around 4 hours and 40 minutes, or 9 hours and 20 minutes of GPU time on H100. Additionally, each epoch takes up roughly the same time.

\subsection{Evaluation}

Moment retrieval is measured on the QVHighlights test set using R1@0.5 (Recall at 1 at 0.5 IoU), R1@0.7 (Recall at 1 at 0.7 IoU), MAP (Mean Average Precision) at 0.5 IoU, MAP at 0.75 IoU, and average MAP. In addition, the highlight detection metric HL MAP and HL HIT@1 are also used. These metrics enable fair comparison with traditional, SFT, and RL models. The test set's answer is not revealed, and official metrics is produced by Codalab website. Due to the limited submission constraints, we did not perform multiple-seed evaluation and variance across runs is not reported.

\section{Results}
\begin{table*}[ht]
\centering
\begin{threeparttable}
\caption{Performance on QVHighlights Test Split of Generative LLM-Based Models}
\label{tab5:qvh_test_set_performance}

\begin{tabular}{llcccccccccc}
    \toprule
    \multirow{2}{*}{\textbf{Method}} 
    & \multirow{2}{*}{\textbf{Base Model}} 
    & \multirow{2}{*}{\textbf{Size}}
    & \multirow{2}{*}{\textbf{Trained}}
    & \multirow{2}{*}{\textbf{Frames}}
    & \multicolumn{2}{c}{\textbf{R1}} 
    & \multicolumn{3}{c}{\textbf{MAP}} 
    & \multicolumn{2}{c}{\textbf{HL}}\\
    \cmidrule(lr){6-7} \cmidrule(lr){8-10} \cmidrule(lr){11-12} 
    &&&&& \textbf{@0.5} & \textbf{@0.7} 
    & \textbf{Avg.} & \textbf{@0.5} & \textbf{@0.75} 
    & \textbf{MAP} & \textbf{HIT@1} \\
    \midrule
    \multicolumn{5}{l}{\textit{\small \textcolor{gray}{Baseline method for QVHighlights: Transformer with custom heads.}}} \\
    Moment-DETR~\cite{qvhighlights_2021} & -- & -- & -- & 75 & 52.89 & 33.02 & 30.73 & 54.82 & 29.40 & \textbf{35.69} & 55.60 \\
    \midrule
    \multicolumn{5}{l}{\textit{\small \textcolor{gray}{Supervised Fine-Tuning methods.}}}\\
    Chrono~\cite{Meinardus_2025_ICCV} & BLIP-2 & 3.94 & 0.02 & 60 & 74.77 & 60.51 & \underline{51.37} & 68.12 & 53.38 & -- & -- \\
    LLaVA-MR~\cite{llava_mr_2024} & BLIP-2 & 3.94 & 0.02 & 80 & 76.59 & 61.48 & -- & \underline{69.41} & \underline{54.40} & -- & -- \\
    VTG-LLM~\cite{vtg_llm_2024} & Video-LLaMA 2 & 8.03 & -- & 96 & -- & -- & -- & -- & -- & 16.50 & 33.50 \\
    VideoChat-T~\cite{zeng2025timesuite} & Mistral & 7.24+ & 7.24+ & 128 & -- & -- & -- & -- & -- & 27.00 & 55.30 \\ 
    SMART~\cite{yu2025smart} & -- & -- & -- & 80 & \underline{78.15} & \underline{63.16} & -- & \textbf{70.76} & \textbf{55.54} & -- & -- \\
    \midrule
    \multicolumn{5}{l}{\textit{\small \textcolor{gray}{Reinforcement Learning methods.}}} \\
    Tempo-R0~\cite{tempo_r0_2025} & Qwen2-VL & 8.30 & 8.30 & -- & \textbf{78.52} & \textbf{65.23} & \textbf{54.50} & -- & -- & -- & -- \\
    TempSamp-R1~\cite{li2025tempsampr1effectivetemporalsampling} & Qwen2.5-VL & 8.30 & 8.30 & 75 & -- & -- & -- & -- & -- & 30.00 & \textbf{57.60} \\ 
    \midrule
    \multicolumn{5}{l}{\textit{\small \textcolor{gray}{SFT \& segmentation method (ours).}}} \\
    BLIP-3 (ours) & BLIP-3 & 4.36 & 0.14 & 25 & 60.77 & 37.42 & 35.28 & 56.26 & 35.80 & \underline{34.48} & \underline{56.74} \\
    \bottomrule
\end{tabular}

\begin{tablenotes}
    \footnotesize
    \item Bold text denotes the best score and underlined text denotes the second best. Empty values mean that they are not reported.
    \item Total sizes taken from HuggingFace are shown in billions for fairness. Trained parameters indicate the number of parameters updated during fine-tuning.
    \item Note that our method uses significantly fewer frames (25) compared to SOTA methods (60+).
    \item Our score is taken on 31 August 2025 from the QVHighlights CodaLab competition page, full logs and score is in our repository.
    
\end{tablenotes}

\end{threeparttable}
\end{table*}

\subsection{Results in Comparison with Other Methods}

Results against other models on the QVHighlights test split are summarized in Table~\ref{tab5:qvh_test_set_performance}. The models displayed in the table were trained on the QVHighlights training split and tested on the test split. To ensure a fair comparison, we explicitly list model sizes, frame counts, and trainable parameters. Note that we only report test set performance because the official labels have not been published, unlike the validation split. 

For moment retrieval, our model achieves higher scores than baseline Moment-DETR~\cite{qvhighlights_2021} on every metric except for HL MAP, but falls short in comparison to SFT/RL based models. However, the majority of the LLM-based models are not able to generate highlights for highlight detection (HD), hence the empty scores. In HD, our model achieves the second-best score for both HL MAP and HIT@1. On the HIT@1 specifically, which evaluates the prediction quality, our method loses only to a larger and more complex RL model (TempSamp-R1~\cite{li2025tempsampr1effectivetemporalsampling}). This confirms that even with coarser temporal sampling (25 frames vs. 60+), the segmentation objective does well in MR and HD. 

\subsection{Training Result}

Offline evaluation using validation split during training shows improvement across epochs (shown in Table~\ref{tab:perf}). The final model from epoch~11 is submitted for online evaluation on the benchmark leaderboard. More importantly, segmentation losses continue to decrease even when the causal LM loss plateaus (Appendix Figure~\ref{fig:training_losses}). This shows that segmentation objectives provide a clear learning signal that is not captured by timestamp-only generation. 

\subsection{Qualitative Results}

\begin{figure*}[ht]
    \centering
    \includegraphics[width=0.85\linewidth]{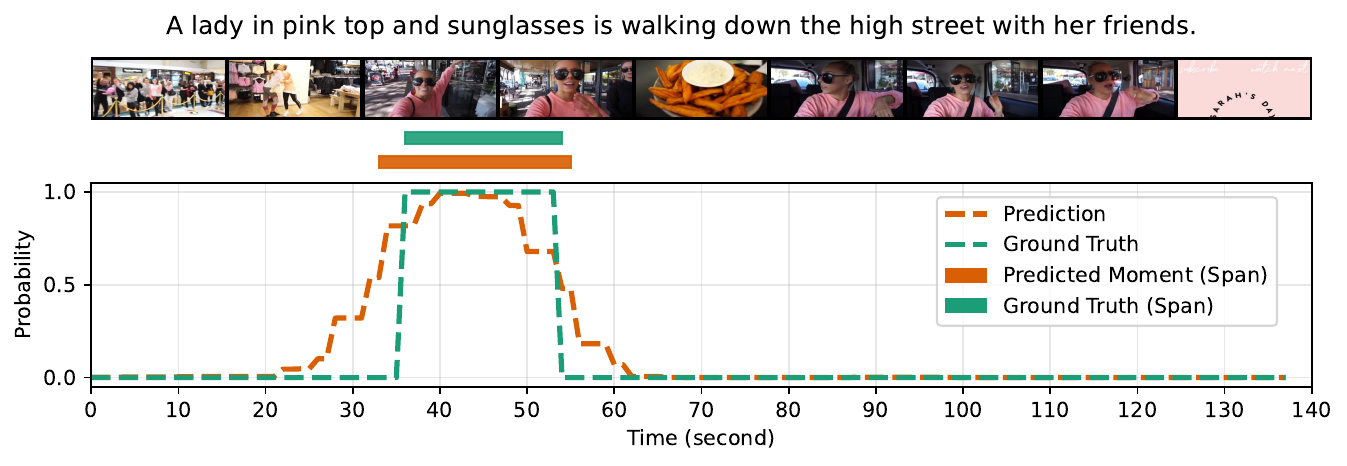}
    \caption{Highlight detection aligns well with ground truth, but moment retrieval is weak. Probabilities remain high ($\ge$ 0.8) over most true segments and only drop after 50s.}
    \label{fig:good_hd}
\end{figure*}

\begin{figure*}[ht]
    \centering
    \includegraphics[width=0.85\linewidth]{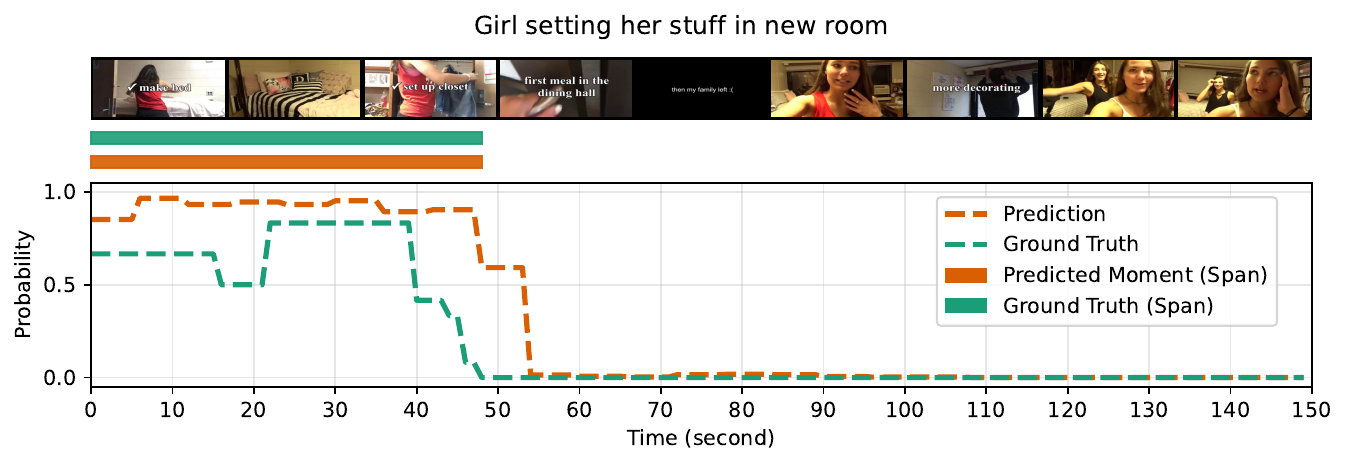}
    \caption{Moment retrieval is perfect, but highlight detection is too inclusive. The model assigns high scores across 0-48s despite true highlights occurring only around 22-39s.}
    \label{fig:good_mr}
\end{figure*}

In this section, we present qualitative examples of two of the most common predictions in Figure~\ref{fig:good_hd} and Figure~\ref{fig:good_mr}. Each figure includes both the predicted and ground-truth in the two tasks: spans for moment retrieval (MR) and probability curve for highlight detection (HD). In most cases, there is a tradeoff between the two tasks. Achieving high IoU for MR and high hit rate for HD at the same time is difficult.
\begin{enumerate}
    \item First, Figure~\ref{fig:good_hd} illustrates a typical case of strong highlight detection but mediocre moment retrieval. The model is able to give high scores on the highlight. The probability curves are similar, and the predicted moment is near.
    \item Conversely, Figure~\ref{fig:good_mr} presents a case where highlight detection fails, yet the model perfectly retrieves the moment. Since the model's output is directly converted to moments, it usually loses span granularity. Perfect moment retrieval like this is quite rare, usually the left/right is slightly misaligned.
\end{enumerate}

\section{Limitations}
The main drawback of this study is the low temporal resolution caused by the 25-frame sampling strategy. While sufficient for validating the use of segmentation objectives, as shown by the strong highlight detection, it caps the IoU for moment retrieval due to the 6-second granularity. Additionally, we selected loss combinations based on dataset characteristics without exhaustive ablation studies. 

Future work, supported by scaled resources, will focus on increasing frame count to 50+ frames, aiming for higher moment retrieval performance. In addition, we will perform ablation study to analyze the contribution of each objective. Furthermore, our method operates on visual frames only, without audio signals. This limitation is evident in cases where visual frames are similar throughout, but audio provides disambiguating information (shown in Figure~\ref{fig:overconfident_prediction}).

\bibliographystyle{IEEEtran}
\bibliography{references}

@ieeetranbstctl{IEEEexample:BSTcontrol,
  ctluse_forced_etal       = {yes},
  ctlmax_names_forced_etal = {6},
  ctlnames_show_etal       = {1},
  ctluse_url               = {no},
  ctluse_doi               = {no},
  ctluse_paper             = {yes}
}

@article{yu2025smart,
  title   = {SMART: Shot-Aware Multimodal Video Moment Retrieval with Audio-Enhanced MLLM},
  author  = {Yu, An and Lu, Weiheng and Li, Jian and Zhang, Zhenfei and Shen, Yunhang and Ye, Felix X-F and Chang, Ming-Ching},
  journal = {arXiv preprint arXiv:2511.14143},
  year    = {2025}
}

@inproceedings{zhai2023sigmoidloss,
  title     = {Sigmoid loss for language image pre-training},
  author    = {Zhai, Xiaohua and Mustafa, Basil and Kolesnikov, Alexander and Beyer, Lucas},
  booktitle = {Proceedings of the IEEE/CVF international conference on computer vision},
  pages     = {11975--11986},
  year      = {2023}
}

@inproceedings{ren2021zero,
  author    = {Jie Ren and Samyam Rajbhandari and Reza Yazdani Aminabadi and Olatunji Ruwase and Shuangyan Yang and Minjia Zhang and Dong Li and Yuxiong He},
  title     = {{ZeRO-Offload}: Democratizing {Billion-Scale} Model Training},
  booktitle = {2021 USENIX Annual Technical Conference (USENIX ATC 21)},
  year      = {2021},
  isbn      = {978-1-939133-23-6},
  pages     = {551--564},
  url       = {https://www.usenix.org/conference/atc21/presentation/ren-jie},
  publisher = {USENIX Association},
  month     = jul
}

@inproceedings{gao2017talltemporalactivitylocalization,
  author    = { Gao, Jiyang and Sun, Chen and Yang, Zhenheng and Nevatia, Ram },
  booktitle = { 2017 IEEE International Conference on Computer Vision (ICCV) },
  title     = {{ TALL: Temporal Activity Localization via Language Query }},
  year      = {2017},
  volume    = {},
  issn      = {2380-7504},
  pages     = {5277-5285},
  doi       = {10.1109/ICCV.2017.563},
  url       = {https://doi.ieeecomputersociety.org/10.1109/ICCV.2017.563},
  publisher = {IEEE Computer Society},
  address   = {Los Alamitos, CA, USA},
  month     = Oct
}

@inproceedings{sun2024tr,
  title     = {Tr-detr: Task-reciprocal transformer for joint moment retrieval and highlight detection},
  author    = {Sun, Hao and Zhou, Mingyao and Chen, Wenjing and Xie, Wei},
  booktitle = {Proceedings of the AAAI Conference on Artificial Intelligence},
  volume    = {38},
  number    = {5},
  pages     = {4998--5007},
  year      = {2024}
}

@inproceedings{zhang2022actionformer,
  title        = {Actionformer: Localizing moments of actions with transformers},
  author       = {Zhang, Chen-Lin and Wu, Jianxin and Li, Yin},
  booktitle    = {European Conference on Computer Vision},
  pages        = {492--510},
  year         = {2022},
  organization = {Springer}
}

@inproceedings{lin2023univtg,
  title     = {Univtg: Towards unified video-language temporal grounding},
  author    = {Lin, Kevin Qinghong and Zhang, Pengchuan and Chen, Joya and Pramanick, Shraman and Gao, Difei and Wang, Alex Jinpeng and Yan, Rui and Shou, Mike Zheng},
  booktitle = {Proceedings of the IEEE/CVF International Conference on Computer Vision},
  pages     = {2794--2804},
  year      = {2023}
}

@inproceedings{moon2023query,
  title     = {Query-dependent video representation for moment retrieval and highlight detection},
  author    = {Moon, WonJun and Hyun, Sangeek and Park, SangUk and Park, Dongchan and Heo, Jae-Pil},
  booktitle = {Proceedings of the IEEE/CVF conference on computer vision and pattern recognition},
  pages     = {23023--23033},
  year      = {2023}
}

@inproceedings{zhao2019hacs,
  title     = {Hacs: Human action clips and segments dataset for recognition and temporal localization},
  author    = {Zhao, Hang and Torralba, Antonio and Torresani, Lorenzo and Yan, Zhicheng},
  booktitle = {Proceedings of the IEEE/CVF International Conference on Computer Vision},
  pages     = {8668--8678},
  year      = {2019}
}

@article{IDREES20171,
  title    = {The THUMOS challenge on action recognition for videos “in the wild”},
  journal  = {Computer Vision and Image Understanding},
  volume   = {155},
  pages    = {1-23},
  year     = {2017},
  issn     = {1077-3142},
  doi      = {https://doi.org/10.1016/j.cviu.2016.10.018},
  url      = {https://www.sciencedirect.com/science/article/pii/S1077314216301710},
  author   = {Haroon Idrees and Amir R. Zamir and Yu-Gang Jiang and Alex Gorban and Ivan Laptev and Rahul Sukthankar and Mubarak Shah}
}

@misc{zhang2025llavavideovideoinstructiontuning,
  title         = {LLaVA-Video: Video Instruction Tuning With Synthetic Data},
  author        = {Yuanhan Zhang and Jinming Wu and Wei Li and Bo Li and Zejun Ma and Ziwei Liu and Chunyuan Li},
  year          = {2025},
  eprint        = {2410.02713},
  archiveprefix = {arXiv},
  primaryclass  = {cs.CV},
  url           = {https://arxiv.org/abs/2410.02713}
}

@inproceedings{zeng2025timesuite,
  title     = {TimeSuite: Improving {MLLM}s for Long Video Understanding via Grounded Tuning},
  author    = {Xiangyu Zeng and Kunchang Li and Chenting Wang and Xinhao Li and Tianxiang Jiang and Ziang Yan and Songze Li and Yansong Shi and Zhengrong Yue and Yi Wang and Yali Wang and Yu Qiao and Limin Wang},
  booktitle = {The Thirteenth International Conference on Learning Representations},
  year      = {2025},
  url       = {https://openreview.net/forum?id=nAVejJURqZ}
}

@misc{rslora_2023,
  title   = {A rank stabilization scaling factor for fine-tuning with lora},
  author  = {Kalajdzievski, Damjan},
  journal = {arXiv preprint arXiv:2312.03732},
  year    = {2023}
}

@inproceedings{Meinardus_2025_ICCV,
  author    = {Meinardus, Boris and Rodriguez, Hector G. and Batra, Anil and Rohrbach, Anna and Rohrbach, Marcus},
  title     = {Chrono: A Simple Blueprint for Representing Time in MLLMs},
  booktitle = {Proceedings of the IEEE/CVF0International0Conference0on0Computer0Vision (ICCV) Workshops},
  month     = {October},
  year      = {2025},
  pages     = {4092-4097}
}

@misc{llava_mr_2024,
  title         = {LLaVA-MR: Large Language-and-Vision Assistant for Video Moment Retrieval},
  author        = {Weiheng Lu and Jian Li and An Yu and Ming-Ching Chang and Shengpeng Ji and Min Xia},
  year          = {2024},
  eprint        = {2411.14505},
  archiveprefix = {arXiv},
  primaryclass  = {cs.CV},
  url           = {https://arxiv.org/abs/2411.14505}
}

@misc{tempo_r0_2025,
  title         = {Tempo-R0: A Video-MLLM for Temporal Video Grounding through Efficient Temporal Sensing Reinforcement Learning},
  author        = {Feng Yue and Zhaoxing Zhang and Junming Jiao and Zhengyu Liang and Shiwen Cao and Feifei Zhang and Rong Shen},
  year          = {2025},
  eprint        = {2507.04702},
  archiveprefix = {arXiv},
  primaryclass  = {cs.CV},
  url           = {https://arxiv.org/abs/2507.04702}
}

@article{vtg_llm_2024,
  title        = {VTG-LLM: Integrating Timestamp Knowledge into Video LLMs for Enhanced Video Temporal Grounding},
  volume       = {39},
  url          = {https://ojs.aaai.org/index.php/AAAI/article/view/32341},
  doi          = {10.1609/aaai.v39i3.32341},
  abstractnote = {Video Temporal Grounding (VTG) strives to accurately pinpoint event timestamps in a specific video using linguistic queries, significantly impacting downstream tasks like video browsing and editing. Unlike traditional task-specific models, Video Large Language Models (video LLMs) can handle multiple tasks concurrently in a zero-shot manner. Consequently, exploring the application of video LLMs for VTG tasks has become a burgeoning research area.
                  However, despite considerable advancements in video content understanding, video LLMs often struggle to accurately pinpoint timestamps within videos, limiting their effectiveness in VTG tasks. To address this, we introduce VTG-LLM, a model designed to enhance video LLMs’ timestamp localization abilities. Our approach includes: (1) effectively integrating timestamp knowledge into visual tokens; (2) incorporating absolute-time tokens to manage timestamp knowledge without concept shifts; and (3) introducing a lightweight, high-performance, slot-based token compression technique designed to accommodate the demands of a large number of frames to be sampled for VTG tasks.
                  Additionally, we present VTG-IT-120K, a collection of publicly available VTG datasets that we have re-annotated to improve upon low-quality annotations.
                  Our comprehensive experiments demonstrate the superior performance of VTG-LLM in comparison to other video LLM methods across a variety of VTG tasks.},
  number       = {3},
  journal      = {Proceedings of the AAAI Conference on Artificial Intelligence},
  author       = {Guo, Yongxin and Liu, Jingyu and Li, Mingda and Cheng, Dingxin and Tang, Xiaoying and Sui, Dianbo and Liu, Qingbin and Chen, Xi and Zhao, Kevin},
  year         = {2025},
  month        = {Apr.},
  pages        = {3302-3310}
}

@inproceedings{liu2022umt,
  title     = {UMT: Unified Multi-modal Transformers for Joint Video Moment Retrieval and Highlight Detection},
  author    = {Liu, Ye and Li, Siyuan and Wu, Yang and Chen, Chang Wen and Shan, Ying and Qie, Xiaohu},
  booktitle = {Proceedings of the IEEE/CVF Conference on Computer Vision and Pattern Recognition (CVPR)},
  pages     = {3042--3051},
  year      = {2022}
}

@inproceedings{7298698,
  author    = {Heilbron, Fabian Caba and Escorcia, Victor and Ghanem, Bernard and Niebles, Juan Carlos},
  booktitle = {2015 IEEE Conference on Computer Vision and Pattern Recognition (CVPR)},
  title     = {ActivityNet: A large-scale video benchmark for human activity understanding},
  year      = {2015},
  volume    = {},
  number    = {},
  pages     = {961-970},
  doi       = {10.1109/CVPR.2015.7298698}
}

@inproceedings{grauman2022ego4d,
  title     = {Ego4d: Around the world in 3,000 hours of egocentric video},
  author    = {Grauman, Kristen and Westbury, Andrew and Byrne, Eugene and Chavis, Zachary and Furnari, Antonino and Girdhar, Rohit and Hamburger, Jackson and Jiang, Hao and Liu, Miao and Liu, Xingyu and others},
  booktitle = {Proceedings of the IEEE/CVF conference on computer vision and pattern recognition},
  pages     = {18995--19012},
  year      = {2022}
}

@misc{li2025tempsampr1effectivetemporalsampling,
  title         = {TempSamp-R1: Effective Temporal Sampling with Reinforcement Fine-Tuning for Video LLMs},
  author        = {Yunheng Li and Jing Cheng and Shaoyong Jia and Hangyi Kuang and Shaohui Jiao and Qibin Hou and Ming-Ming Cheng},
  year          = {2025},
  eprint        = {2509.18056},
  archiveprefix = {arXiv},
  primaryclass  = {cs.CV},
  url           = {https://arxiv.org/abs/2509.18056}
}

@misc{abdin2024phi3technicalreporthighly,
  title         = {Phi-3 Technical Report: A Highly Capable Language Model Locally on Your Phone},
  author        = {Marah Abdin and Jyoti Aneja and Hany Awadalla and Ahmed Awadallah and Ammar Ahmad Awan and Nguyen Bach and Amit Bahree and Arash Bakhtiari and Jianmin Bao and Harkirat Behl and Alon Benhaim and Misha Bilenko and Johan Bjorck and Sébastien Bubeck and Martin Cai and Qin Cai and Vishrav Chaudhary and Dong Chen and Dongdong Chen and Weizhu Chen and Yen-Chun Chen and Yi-Ling Chen and Hao Cheng and Parul Chopra and Xiyang Dai and Matthew Dixon and Ronen Eldan and Victor Fragoso and Jianfeng Gao and Mei Gao and Min Gao and Amit Garg and Allie Del Giorno and Abhishek Goswami and Suriya Gunasekar and Emman Haider and Junheng Hao and Russell J. Hewett and Wenxiang Hu and Jamie Huynh and Dan Iter and Sam Ade Jacobs and Mojan Javaheripi and Xin Jin and Nikos Karampatziakis and Piero Kauffmann and Mahoud Khademi and Dongwoo Kim and Young Jin Kim and Lev Kurilenko and James R. Lee and Yin Tat Lee and Yuanzhi Li and Yunsheng Li and Chen Liang and Lars Liden and Xihui Lin and Zeqi Lin and Ce Liu and Liyuan Liu and Mengchen Liu and Weishung Liu and Xiaodong Liu and Chong Luo and Piyush Madan and Ali Mahmoudzadeh and David Majercak and Matt Mazzola and Caio César Teodoro Mendes and Arindam Mitra and Hardik Modi and Anh Nguyen and Brandon Norick and Barun Patra and Daniel Perez-Becker and Thomas Portet and Reid Pryzant and Heyang Qin and Marko Radmilac and Liliang Ren and Gustavo de Rosa and Corby Rosset and Sambudha Roy and Olatunji Ruwase and Olli Saarikivi and Amin Saied and Adil Salim and Michael Santacroce and Shital Shah and Ning Shang and Hiteshi Sharma and Yelong Shen and Swadheen Shukla and Xia Song and Masahiro Tanaka and Andrea Tupini and Praneetha Vaddamanu and Chunyu Wang and Guanhua Wang and Lijuan Wang and Shuohang Wang and Xin Wang and Yu Wang and Rachel Ward and Wen Wen and Philipp Witte and Haiping Wu and Xiaoxia Wu and Michael Wyatt and Bin Xiao and Can Xu and Jiahang Xu and Weijian Xu and Jilong Xue and Sonali Yadav and Fan Yang and Jianwei Yang and Yifan Yang and Ziyi Yang and Donghan Yu and Lu Yuan and Chenruidong Zhang and Cyril Zhang and Jianwen Zhang and Li Lyna Zhang and Yi Zhang and Yue Zhang and Yunan Zhang and Xiren Zhou},
  year          = {2024},
  eprint        = {2404.14219},
  archiveprefix = {arXiv},
  primaryclass  = {cs.CL},
  url           = {https://arxiv.org/abs/2404.14219}
}

@article{alayrac2022flamingo,
  title   = {Flamingo: a visual language model for few-shot learning},
  author  = {Alayrac, Jean-Baptiste and Donahue, Jeff and Luc, Pauline and Miech, Antoine and Barr, Iain and Hasson, Yana and Lenc, Karel and Mensch, Arthur and Millican, Katherine and Reynolds, Malcolm and others},
  journal = {Advances in neural information processing systems},
  volume  = {35},
  pages   = {23716--23736},
  year    = {2022}
}

@inproceedings{qvhighlights_2021,
  author    = {Lei, Jie and Berg, Tamara L and Bansal, Mohit},
  booktitle = {Advances in Neural Information Processing Systems},
  editor    = {M. Ranzato and A. Beygelzimer and Y. Dauphin and P.S. Liang and J. Wortman Vaughan},
  pages     = {11846--11858},
  publisher = {Curran Associates, Inc.},
  title     = {Detecting Moments and Highlights in Videos via Natural Language Queries},
  url       = {https://proceedings.neurips.cc/paper_files/paper/2021/file/62e0973455fd26eb03e91d5741a4a3bb-Paper.pdf},
  volume    = {34},
  year      = {2021}
}

@inproceedings{lisa_2024,
  author    = {Lai, Xin and Tian, Zhuotao and Chen, Yukang and Li, Yanwei and Yuan, Yuhui and Liu, Shu and Jia, Jiaya},
  title     = {LISA: Reasoning Segmentation via Large Language Model},
  booktitle = {Proceedings of the IEEE/CVF Conference on Computer Vision and Pattern Recognition (CVPR)},
  month     = {June},
  year      = {2024},
  pages     = {9579-9589}
}

@article{vaswani2017attention,
  title   = {Attention is all you need},
  author  = {Vaswani, Ashish and Shazeer, Noam and Parmar, Niki and Uszkoreit, Jakob and Jones, Llion and Gomez, Aidan N and Kaiser, {\L}ukasz and Polosukhin, Illia},
  journal = {Advances in neural information processing systems},
  volume  = {30},
  year    = {2017}
}

@inproceedings{abraham2019focaltversky,
  author    = {Abraham, Nabila and Khan, Naimul Mefraz},
  booktitle = {2019 IEEE 16th International Symposium on Biomedical Imaging (ISBI 2019)},
  title     = {A Novel Focal Tversky Loss Function With Improved Attention U-Net for Lesion Segmentation},
  year      = {2019},
  volume    = {},
  number    = {},
  pages     = {683-687},
  doi       = {10.1109/ISBI.2019.8759329}
}

@inproceedings{qian2025reasoning,
  title     = {Reasoning to attend: Try to understand how< seg> token works},
  author    = {Qian, Rui and Yin, Xin and Dou, Dejing},
  booktitle = {Proceedings of the Computer Vision and Pattern Recognition Conference},
  pages     = {24722--24731},
  year      = {2025}
}

@article{taghanaki2019combo,
  title     = {Combo loss: Handling input and output imbalance in multi-organ segmentation},
  author    = {Taghanaki, Saeid Asgari and Zheng, Yefeng and Zhou, S Kevin and Georgescu, Bogdan and Sharma, Puneet and Xu, Daguang and Comaniciu, Dorin and Hamarneh, Ghassan},
  journal   = {Computerized Medical Imaging and Graphics},
  volume    = {75},
  pages     = {24--33},
  year      = {2019},
  publisher = {Elsevier}
}

@article{wu2024visionllm,
  title   = {Visionllm v2: An end-to-end generalist multimodal large language model for hundreds of vision-language tasks},
  author  = {Wu, Jiannan and Zhong, Muyan and Xing, Sen and Lai, Zeqiang and Liu, Zhaoyang and Chen, Zhe and Wang, Wenhai and Zhu, Xizhou and Lu, Lewei and Lu, Tong and others},
  journal = {Advances in Neural Information Processing Systems},
  volume  = {37},
  pages   = {69925--69975},
  year    = {2024}
}

@misc{azad2023loss,
  title         = {Loss Functions in the Era of Semantic Segmentation: A Survey and Outlook},
  author        = {Reza Azad and Moein Heidary and Kadir Yilmaz and Michael Hüttemann and Sanaz Karimijafarbigloo and Yuli Wu and Anke Schmeink and Dorit Merhof},
  year          = {2023},
  eprint        = {2312.05391},
  archiveprefix = {arXiv},
  primaryclass  = {cs.CV}
}

@article{liu2025seg,
  title   = {Seg-zero: Reasoning-chain guided segmentation via cognitive reinforcement},
  author  = {Liu, Yuqi and Peng, Bohao and Zhong, Zhisheng and Yue, Zihao and Lu, Fanbin and Yu, Bei and Jia, Jiaya},
  journal = {arXiv preprint arXiv:2503.06520},
  year    = {2025}
}

@misc{lan2024text4seg,
  title         = {Text4Seg: Reimagining Image Segmentation as Text Generation},
  author        = {Mengcheng Lan and Chaofeng Chen and Yue Zhou and Jiaxing Xu and Yiping Ke and Xinjiang Wang and Litong Feng and Wayne Zhang},
  year          = {2024},
  eprint        = {2410.09855},
  archiveprefix = {arXiv},
  primaryclass  = {cs.CV},
  url           = {https://arxiv.org/abs/2410.09855}
}

@inproceedings{jang2025mmr,
  title     = {{MMR}: A Large-scale Benchmark Dataset for Multi-target and Multi-granularity Reasoning Segmentation},
  author    = {Donggon Jang and Yucheol Cho and Suin Lee and Taehyeon Kim and Daeshik Kim},
  booktitle = {The Thirteenth International Conference on Learning Representations},
  year      = {2025},
  url       = {https://openreview.net/forum?id=mzL19kKE3r}
}

@misc{blip3_2025,
  title         = {xGen-MM (BLIP-3): A Family of Open Large Multimodal Models},
  author        = {Le Xue and Manli Shu and Anas Awadalla and Jun Wang and An Yan and Senthil Purushwalkam and Honglu Zhou and Viraj Prabhu and Yutong Dai and Michael S Ryoo and Shrikant Kendre and Jieyu Zhang and Shaoyen Tseng and Gustavo A Lujan-Moreno and Matthew L Olson and Musashi Hinck and David Cobbley and Vasudev Lal and Can Qin and Shu Zhang and Chia-Chih Chen and Ning Yu and Juntao Tan and Tulika Manoj Awalgaonkar and Shelby Heinecke and Huan Wang and Yejin Choi and Ludwig Schmidt and Zeyuan Chen and Silvio Savarese and Juan Carlos Niebles and Caiming Xiong and Ran Xu},
  year          = {2025},
  eprint        = {2408.08872},
  archiveprefix = {arXiv},
  primaryclass  = {cs.CV},
  url           = {https://arxiv.org/abs/2408.08872}
}

@inproceedings{pissa_2024,
  author    = {Meng, Fanxu and Wang, Zhaohui and Zhang, Muhan},
  booktitle = {Advances in Neural Information Processing Systems},
  editor    = {A. Globerson and L. Mackey and D. Belgrave and A. Fan and U. Paquet and J. Tomczak and C. Zhang},
  pages     = {121038--121072},
  publisher = {Curran Associates, Inc.},
  title     = {PiSSA: Principal Singular Values and Singular Vectors Adaptation of Large Language Models},
  url       = {https://proceedings.neurips.cc/paper_files/paper/2024/file/db36f4d603cc9e3a2a5e10b93e6428f2-Paper-Conference.pdf},
  volume    = {37},
  year      = {2024}
}

@inproceedings{generalized_dice_2017,
  author    = {Sudre, Carole H.
               and Li, Wenqi
               and Vercauteren, Tom
               and Ourselin, Sebastien
               and Jorge Cardoso, M.},
  editor    = {Cardoso, M. Jorge
               and Arbel, Tal
               and Carneiro, Gustavo
               and Syeda-Mahmood, Tanveer
               and Tavares, Jo{\~a}o Manuel R.S.
               and Moradi, Mehdi
               and Bradley, Andrew
               and Greenspan, Hayit
               and Papa, Jo{\~a}o Paulo
               and Madabhushi, Anant
               and Nascimento, Jacinto C.
               and Cardoso, Jaime S.
               and Belagiannis, Vasileios
               and Lu, Zhi},
  title     = {Generalised Dice Overlap as a Deep Learning Loss Function for Highly Unbalanced Segmentations},
  booktitle = {Deep Learning in Medical Image Analysis and Multimodal Learning for Clinical Decision Support },
  year      = {2017},
  publisher = {Springer International Publishing},
  address   = {Cham},
  pages     = {240--248},
  isbn      = {978-3-319-67558-9}
}

@inproceedings{rezatofighi2019generalized,
  title     = {Generalized intersection over union: A metric and a loss for bounding box regression},
  author    = {Rezatofighi, Hamid and Tsoi, Nathan and Gwak, JunYoung and Sadeghian, Amir and Reid, Ian and Savarese, Silvio},
  booktitle = {Proceedings of the IEEE/CVF conference on computer vision and pattern recognition},
  pages     = {658--666},
  year      = {2019}
}

@misc{shao2024deepseekmathpushinglimitsmathematical,
  title         = {DeepSeekMath: Pushing the Limits of Mathematical Reasoning in Open Language Models},
  author        = {Zhihong Shao and Peiyi Wang and Qihao Zhu and Runxin Xu and Junxiao Song and Xiao Bi and Haowei Zhang and Mingchuan Zhang and Y. K. Li and Y. Wu and Daya Guo},
  year          = {2024},
  eprint        = {2402.03300},
  archiveprefix = {arXiv},
  primaryclass  = {cs.CL},
  url           = {https://arxiv.org/abs/2402.03300}
}

@inproceedings{tversky_2017,
  author    = {Salehi, Seyed Sadegh Mohseni
               and Erdogmus, Deniz
               and Gholipour, Ali},
  editor    = {Wang, Qian
               and Shi, Yinghuan
               and Suk, Heung-Il
               and Suzuki, Kenji},
  title     = {Tversky Loss Function for Image Segmentation Using 3D Fully Convolutional Deep Networks},
  booktitle = {Machine Learning in Medical Imaging},
  year      = {2017},
  publisher = {Springer International Publishing},
  address   = {Cham},
  pages     = {379--387},
  isbn      = {978-3-319-67389-9}
}

\clearpage
\appendices
\section{Prompt Design}
\noindent Below is the design of the question prompt for 25 frames. Frames 1-23 are omitted for concision. The \texttt{<image>} string are placeholders for the vision tokens and is decoded to a single special token by the BLIP-3 tokenizer.
\begin{lstlisting}[breaklines=true, breakindent=0pt]
You are given 25 frames sampled from a video, ordered and separated by newline characters, indexed from 0 to 24:
Frame 0: <image>
... 
Frame 24: <image>

**Your task**: given an activity, analyze the video frames to identify which ones contain the specified activity.

**Output format**: provide a 25 character binary segmentation mask, specifically:
- '1' means the frame at that position likely matches to the activity.
- '0' means the frame at that position likely does not match to the activity.
- Your output must be exactly 25 characters long and contain only '1's and '0's, with no spaces or other delimiters and no explanations.

**Activity**: Man in baseball cap eats before doing his interview.

**Question**: Which frames contains the activity?
\end{lstlisting}

\noindent Here is the answer example for the prompt.
\begin{lstlisting}[breaklines=true]
0000000000001111111111010
\end{lstlisting}

\noindent For both training and evaluation, the following system prompt is used.
\begin{lstlisting}[breaklines=true, breakindent=0pt]
You are a smart video retrieval assistant. 
You will receive a video and a human activity query given by the user. 
Return the frames that matches the activity query. 
Follow the output format given by the user.
\end{lstlisting}

The prompt is heavily inspired by synthetic data generation with GPT-4o from LLaVA-Video~\cite{zhang2025llavavideovideoinstructiontuning} where detailed instruction with markdown-like syntax is used. Particularly, lists with dashes and bold texts (text in between two double asterisks). 

\section{Preprocessing Details}
\subsection{Frame Processing}
At training/inference, each frame is resized to SigLIP~\cite{zhai2023sigmoidloss} (specifically, the "siglip-so400m-patch14-384" variant) input dimension, which is $384\times 384$ ($h\times w$). Then, each RGB channel is normalized to 0.5 mean and 0.5 std, and forwarded into the SigLIP model. SigLIP then outputs a visual features $\mathbf{O}\in{\mathbb{R}^{f\times{729}\times d_{siglip}}}$ which is then flattened to $\mathbf{O}^\prime\in{\mathbb{R}^{(729f)\times d_{siglip}}}$.

Following the Perceiver Resampler details at \cite{alayrac2022flamingo}, flattened output $\mathbf{O}^\prime$ is then forwarded to the Perceiver Resampler. During this forward pass, a set of learned latent queries $\mathbf{X}\in{\mathbb{R}^{R\times{d_{LLM}}}}$ is used as query matrix while $concat([\mathbf{O}^\prime, \mathbf{X}])$ is used as key and value matrix for a standard attention mechanism~\cite{vaswani2017attention}. Output of the Perceiver Resampler matches the embedding dimension used by the LLM. It is then interleaved to the prompt, replacing the \texttt{<image>} tokens, and fed into the LLM. 

\subsection{Frame Sampling}
\begin{figure}[h]
    \centering
    \includegraphics[width=0.8\linewidth]{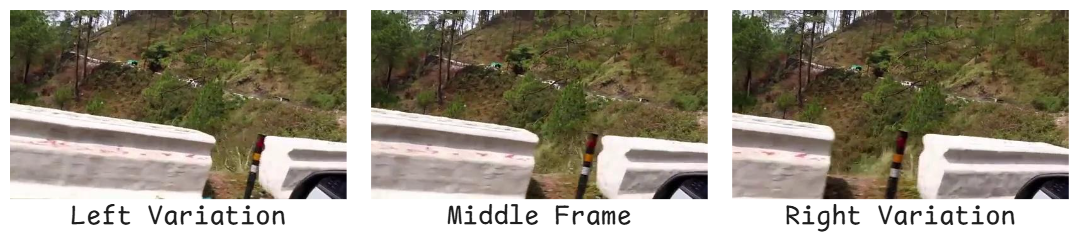}
    \caption{Frame sampling example. For a fast-moving frame like this, the variations might differ a lot, but the effect is minimal on a slow-moving frame.}
    \label{fig:frame_sampling}
\end{figure}

The variation sampling can be seen in Figure~\ref{fig:frame_sampling}. The sampling is done at each frame and is randomly picked at training to increase the amount of data. Because the dataset contains mostly videos with 30 FPS, The left and right variations are exactly 3 frames apart from the middle frame.

\section{Hyperparameter Details}
The hyperparameters used in model training can be seen in Table~\ref{tab:hyperparams}. The training is done with Python 3.12 and PyTorch 2.7.1. The loss weights are also explained in Table~\ref{tab:loss_weights}. Additionally, the model is trained on 2 H100 GPUs with CUDA version 12.8. To construct efficient and fast multi-GPU training, DeepSpeed is used and the configuration Zero-2 Offload is picked \cite{ren2021zero}.

\begin{table}[htbp]
\caption{Training Hyperparameters}
\label{tab:hyperparams}
\centering
\begin{threeparttable}
\begin{tabular}{ll}
\toprule
\textbf{Hyperparameter} & \textbf{Value} \\
\midrule
    Learning rate & $2e^{-5}$ \\
    Gradient accumulation & 4 \\
    Batch size per GPU & 16 \\
    PiSSA rank & 16 \\
    % PiSSA alpha & 16 \\
    PiSSA iteration & 4 \\
    PiSSA dropout rate & 0.025 \\
    Weight decay & 0.005 \\
    Tversky loss $\beta$ & 0.7 \\
    BCE positive weight & 2.3378 \\
    Warm-up epochs & 6 \\
    LR scheduler & Cosine decay \\
    DeepSpeed & Zero-2 Offload \\
    Number of frames & 25 \\
    % Precision & Bfloat 16 \\
    Number of beams & 2 \\
    Number of epochs & 11 \\
    % Seed & 2109 \\
\bottomrule
\end{tabular}
\begin{tablenotes}
    \footnotesize
    \item Same weight decay is used for the language model and vision tokenizer
    % \item Seed is set for these Python libraries: Numpy, PyTorch, and Random (standard library). 
\end{tablenotes}
\end{threeparttable}
\end{table}

\begin{table}[h]
\caption{Loss Weights during Training}
\label{tab:loss_weights}
\centering
\begin{threeparttable}
    
\begin{tabular}{l|cc}
    \toprule
    \textbf{Loss Function} & \textbf{Epoch 1} & \textbf{Epoch 7}\\
    \midrule
    Cross-entropy loss ($w_{lm}$) & 1.0 & 0.2000 \\
    Binary cross-entropy loss ($w_{bce}$) & 0.0 & 0.2667 \\
    Tversky loss ($w_{tv}$) & 0.0 & 0.2667 \\
    Generalized Dice loss ($w_{gd}$) & 0.0 & 0.2667 \\
\end{tabular}
\begin{tablenotes}
    \footnotesize
    \item The model is warmed-up for 6 epochs and all losses are increased or decreased linearly.
\end{tablenotes}
\end{threeparttable}
\end{table}

\section{Improvements over Epochs}
\begin{figure*}[htbp]
    \centering
    \includegraphics[width=0.8\linewidth]{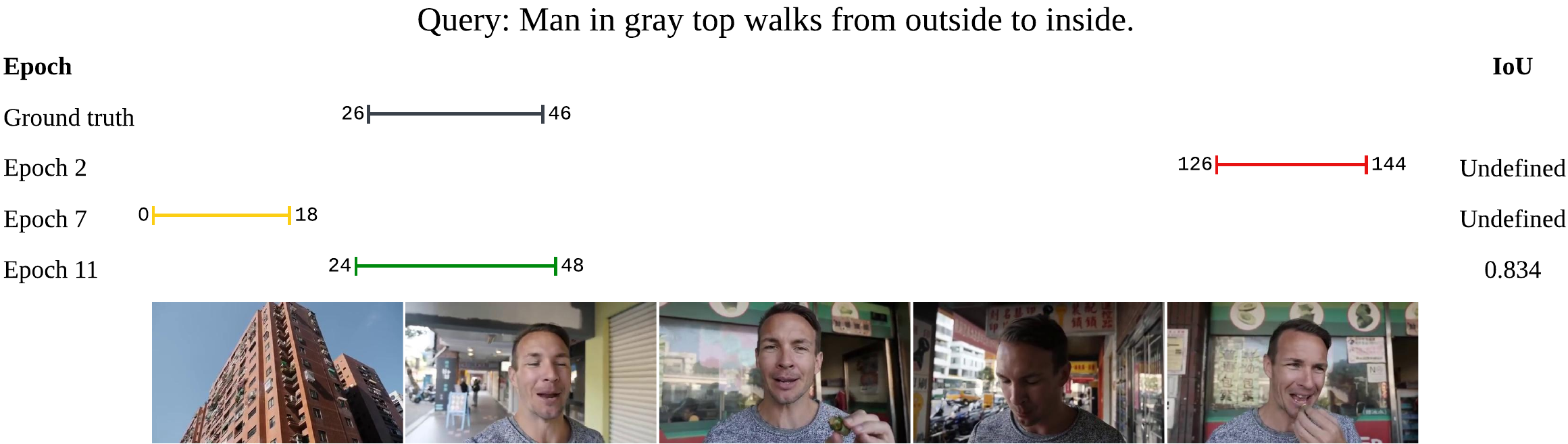}
    \caption{Example 1 of model improvement from early (epoch 2) to final (epoch 11) training. At the final epoch, model is able to pinpoint the moment with great IoU.}
    \label{fig:d1_example1}
\end{figure*}

\begin{figure*}[htbp]
    \centering
    \includegraphics[width=0.8\linewidth]{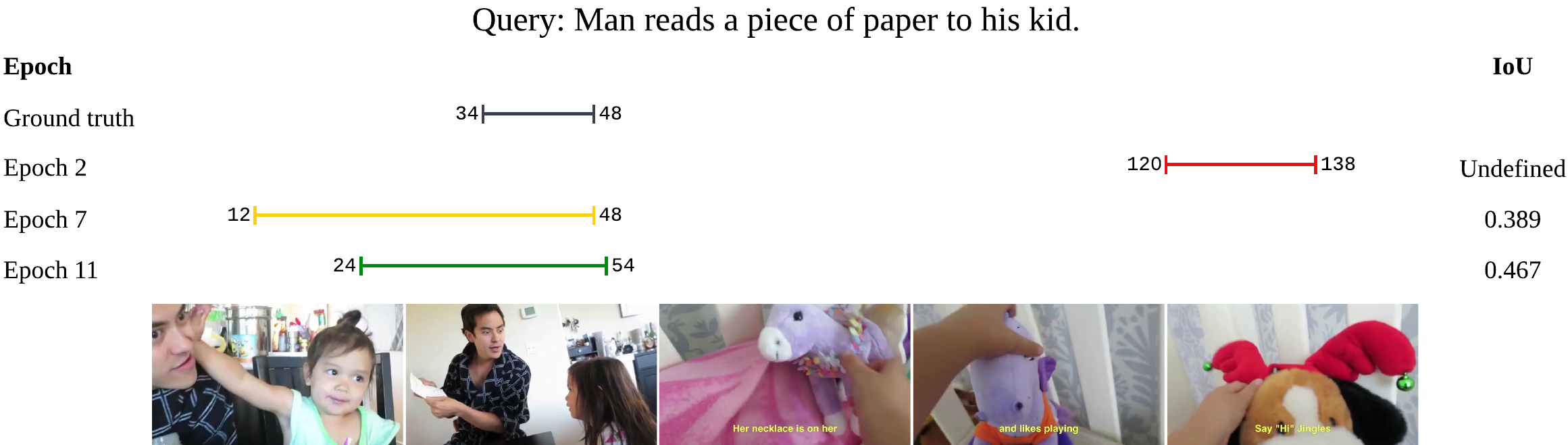}
    \caption{Example 2 of model improvement from early (epoch 2) to final (epoch 11) training. While the 7\textsuperscript{th} epoch variant already encloses the ground truth, signals from later epochs improve the model, resulting in a tighter span.}
    \label{fig:d1_example2}
\end{figure*}

We report examples of the learning process in Figure~\ref{fig:d1_example1} and Figure~\ref{fig:d1_example2}. At epoch 2, the model almost always assign high scores at around the end of the video. While not every sample is displayed, this pattern is quite consistent. In the later epochs, the model is able to adjust to find the correct moment. 

\section{More Qualitative Results}
\begin{figure*}[htbp]
    \centering
    \includegraphics[width=0.8\linewidth]{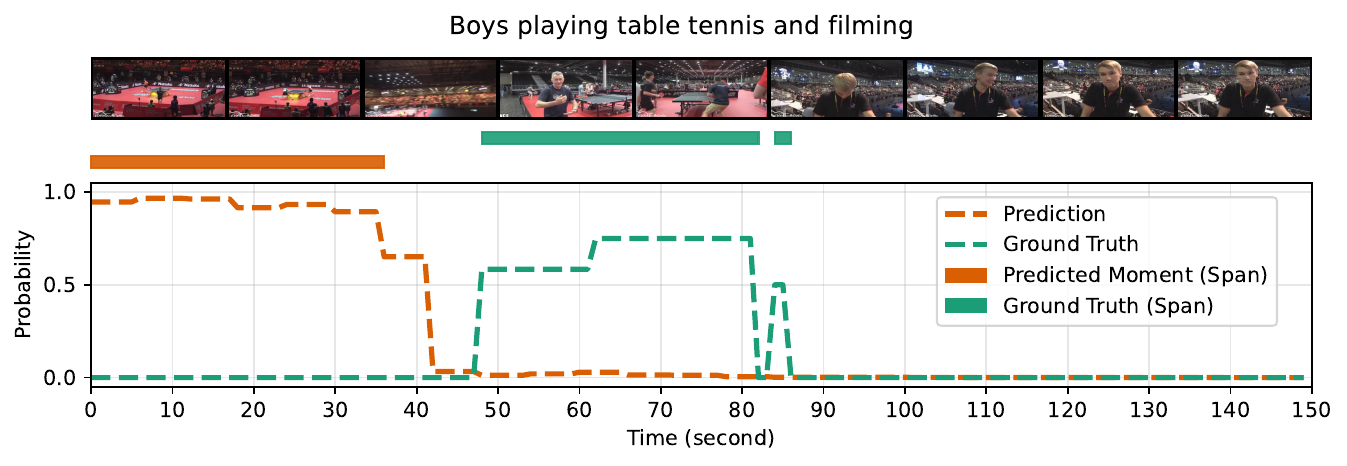}
    \caption{The probability peak aligns with the vlogger's casual play instead of the athletes' matches. The model fails understand that unlike the athletes, the boys are playing while filming at the same time (vlogging).}
    \label{fig:wrong_prediction}
\end{figure*}

\begin{figure*}[htbp]
    \centering
    \includegraphics[width=0.8\linewidth]{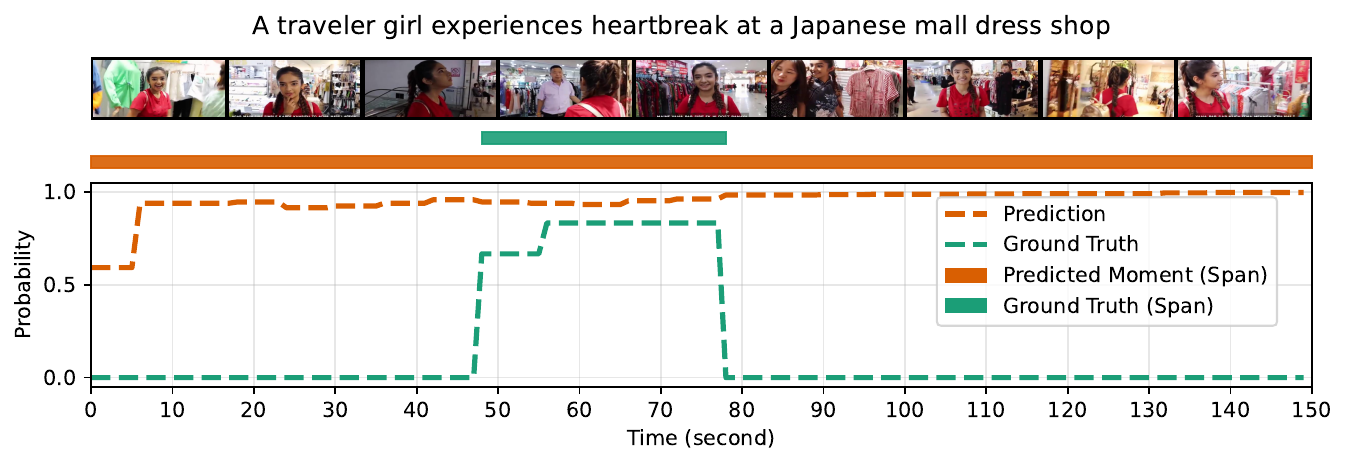}
    \caption{Overconfident prediction where model thinks all the frames are related to the query. In this case, the lack of audio input confuses the model since every frame is similar.}
    \label{fig:overconfident_prediction}
\end{figure*}

\begin{figure*}[htbp]
    \centering
    \includegraphics[width=0.8\linewidth]{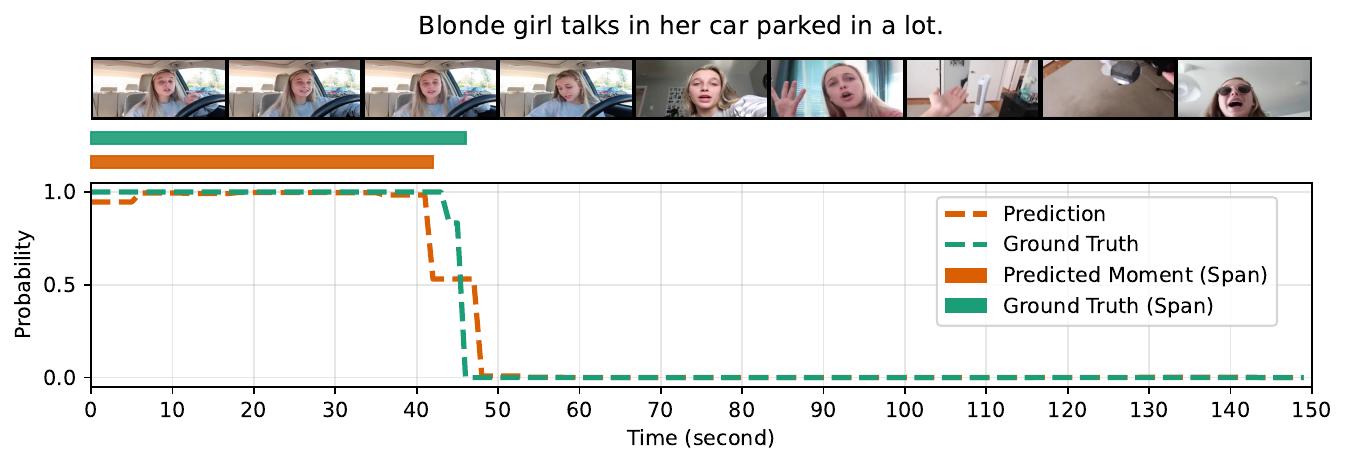}
    \caption{Excellent span and probability prediction. The model correctly identifies the girl's speaking segments and assigns low probability on the non-speech portions, even when the girl is still in the car.}
    \label{fig:best}
\end{figure*}

In addition to section V.C, we present other examples that commonly appear in prediction. 
\begin{enumerate}
    \item In Figure~\ref{fig:wrong_prediction}, we show a failure case. The probability curve peaks in the wrong place and is very low in the right place.  
    \item Figure~\ref{fig:overconfident_prediction} displays a typical overconfident prediction. The model predicts all the frames to be relevant though only a small portion actually is. While these kinds of predictions have non-zero recall, MR metrics always have a threshold that excludes these predictions from the global statistics.
    \item An excellent prediction is seen at Figure~\ref{fig:best} where not only the moments are very similar, but the predicted probabilities are also aligned very well.
\end{enumerate}

\section{Additional Results}
\begin{figure*}[htbp]
    \centering
    \includegraphics[width=0.8\linewidth]{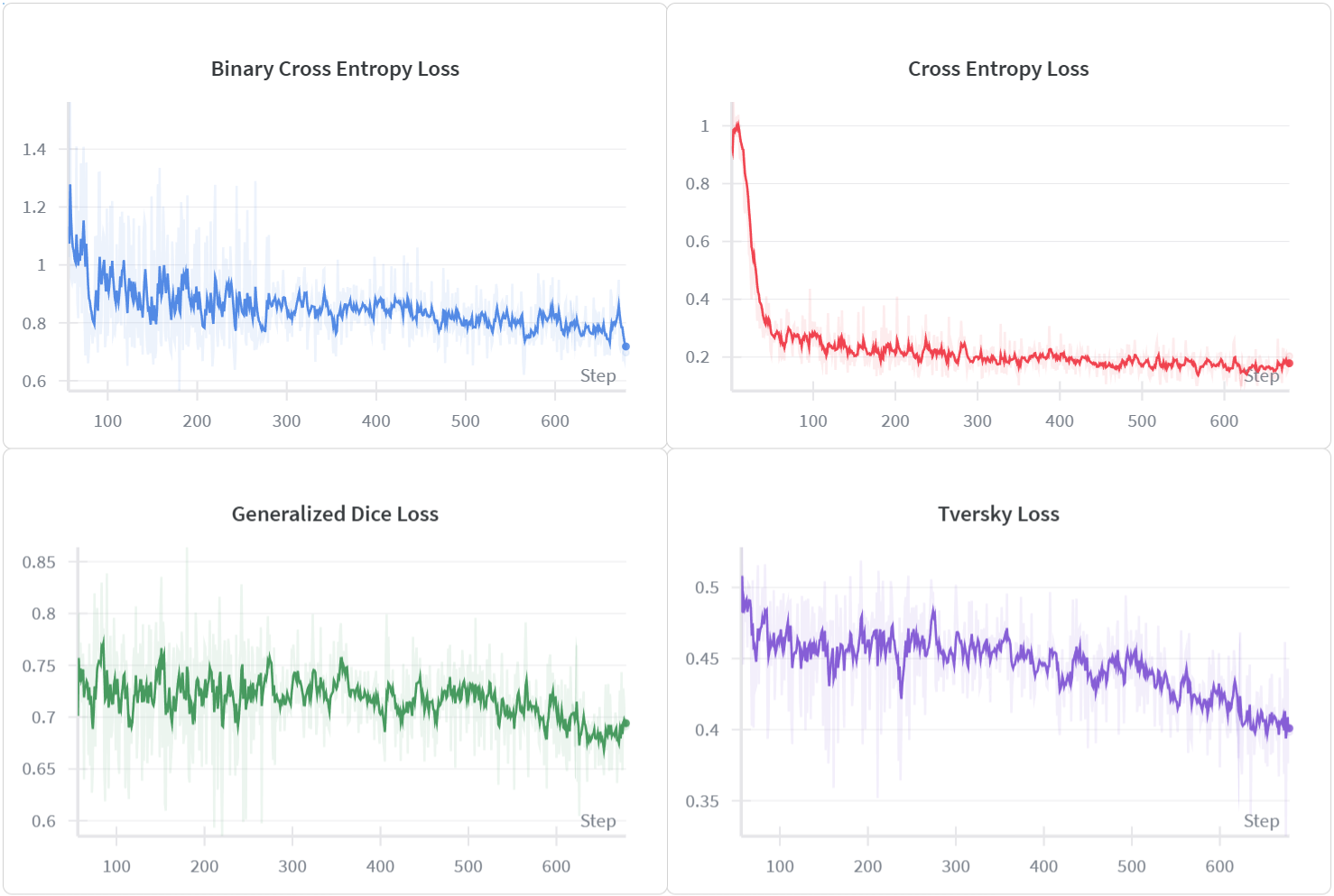}
    \caption{Loss functions during training. The charts show each loss component and their value over time. To enhance readability, time weighted EMA smoothing is applied to each curve, overlaying the original values at lower opacity.}
    \label{fig:training_losses}
\end{figure*}

This section presents additional training curves for the four losses to support the results, shown in Figure~\ref{fig:training_losses}. From the three segmentation losses, Tversky loss converges the best, showing a clear downtrend. While generalized dice and binary cross-entropy also decreased over time, the trend is not as prominent. In addition, the performance per epoch (measured offline) is reported in Table~\ref{tab:perf}. Massive performance gains can be seen in epoch 5 and epoch 9.

\begin{table}[htbp]
\caption{Performance per epoch on QVHighlights validation set}
\label{tab:perf}
    \centering
    \begin{threeparttable}
    \begin{tabular}{c|ccc}
        \toprule
        \textbf{Epoch} & \textbf{R1@0.7} & \textbf{Avg. MAP} & \textbf{MAP@0.75} \\
        \midrule
        1 & 2.22 & 2.50 & 1.89 \\
        2 & 13.28 & 7.61 & 6.98 \\
        3 & 11.80 & 10.87 & 9.41 \\
        4 & 17.78 & 15.65 & 14.04 \\
        5 & 30.14 & 24.53 & 24.03 \\
        6 & 35.61 & 29.41 & 29.49 \\
        7 & 37.86 & 30.71 & 30.65 \\
        8 & 36.62 & 29.96 & 29.96 \\
        9 & 43.15 & 33.99 & 35.09 \\
        10 & 42.70 & 33.72 & 34.06 \\
        11 & 43.74 & 35.01 & 35.90 \\
        \bottomrule
    \end{tabular}
    \begin{tablenotes}
        \footnotesize
        \item The scores are single-run scores achieved on offline evaluation
    \end{tablenotes}
    \end{threeparttable}
\end{table}

\end{document}